\documentclass[11pt,a4paper]{article}
\usepackage{acl2019}
\usepackage{times}
\usepackage{latexsym}

\usepackage{url}

\usepackage{bm}
\usepackage{amsmath}
\usepackage{amssymb}
\usepackage{multirow}
\usepackage{graphicx}
\usepackage{subfigure}
\usepackage[T1]{fontenc}
\usepackage[utf8]{inputenc}
\usepackage{CJKutf8}
\usepackage{lineno,hyperref}
\usepackage{url}
\usepackage{enumitem}
\usepackage{amsfonts,amssymb}
\usepackage[ruled]{algorithm2e}
\usepackage{fancyhdr}

\aclfinalcopy 

\title{Controllable Data Synthesis Method for Grammatical Error Correction}

\author{Liner Yang$^\S$, Chencheng Wang$^\dagger$$^\S$,  Yun Chen$^\ddagger$, Yongping Du$^\dagger$,  Erhong Yang$^\S$\\
	$^\S$Beijing Language and Culture University, Beijing, China  \\
	$^\dagger$Beijing University of Technology, Beijing, China \\
	$^\ddagger$Shanghai University of Finance and Economics, Shanghai, China \\
{\tt \{lineryang,hsamswang,yun.chencreek\}@gmail.com} \\
{\tt ypdu@bjut.edu.cn, yerhong@blcu.edu.cn}
}

\date{}
\newcommand{\makeheadrule}{%
	\makebox[0pt][l]{\rule[0.0\baselineskip]{\headwidth}{0.0pt}}%
	\rule[0.0\baselineskip]{\headwidth}{0.0pt}}
\renewcommand{\headrule}{%
	{\if@fancyplain\let\headrulewidth\plainheadrulewidth\fi
		\makeheadrule}}
\makeatother
\begin{document}
\maketitle
\begin{abstract}
  Due to the lack of parallel data in current Grammatical Error Correction (GEC) task, models based on Sequence to Sequence framework cannot be adequately trained to obtain higher performance. We propose two data synthesis methods which can control the error rate and the ratio of error types on synthetic data. The first approach is to corrupt each word in the monolingual corpus with a fixed probability, including replacement, insertion and deletion. Another approach is to train error generation models and further filtering the decoding results of the models. The experiments on different synthetic data show that the error rate is 40\% and the ratio of error types is the same can improve the model performance better. Finally, we synthesize about 100 million data and achieve comparable performance as the state of the art, which uses twice as much data as we use.
\end{abstract}

\begin{CJK*}{UTF8}{gbsn}
\section{Introduction}
The task of Grammatical Error Correction is primarily aimed at detecting and correcting errors in essays. e.g. I follows his advices $\rightarrow$ I followed his advice. With the increasing number of English as a Second Language learners (ESL) and the success of four shared tasks \cite{HOO2011,HOO2012,CoNLL2013,CoNLL2014}, more and more researchers focus on the study of GEC.

Since Brockett\cite{TheFirstMTGEC} introduced the idea of translation to solve the problem of GEC by translating error text into right one, a lot of sequence to sequence models have been applied to GEC \cite{NUS2018CNN,NUS2018Quality,JD2018a,ge2018fluency}. All of these studies are based on two public corpora, the largest parallel corpus Lang-8 \cite{Lang-8} and NUS Corpus of Learner English (NUCLE) \cite{NUCLEcorpus}.

Although these corpora contain millions of sentence pairs, the model still cannot be trained to achieve higher performance in GEC tasks. Thus, the state-of-the-art systems \cite{JD2018b,yuanfudao,google2019} regard GEC as a low-resource Neural Machine Translation (NMT) task. Neither of their methods takes full account of the impact of different error rates and error type ratios on the results. Therefore, we propose two data synthesis methods that can control the error information of synthetic data. The effects of different error rates, error types and the size of the synthesized corpus on the performance of the model are also discussed.

The first approach is rule-based, corrupting each token of monolingual corpus with a fixed probability according to the required error rate. Corruption operations include deleting tokens, inserting and replacing random tokens in the vocabulary. The ratio of error types on the synthetic data is controlled by the probability of operations.

The second approach is based on Back translation mechanism. We separately train statistical machine translation (SMT) and NMT models on the corpora of $ correct\rightarrow error$ sentence pairs and introduce noise to the monolingual corpus in the decoding stage. In order to control the error distribution of synthetic data, we propose a filtering strategy to remove instances with lower error rate and types of errors that we don't need. 

By controlling variables, we explore the effects of different error rates and ratio of error types on the model performance. Finally, we pretrain our model on 100 million synthesized data and achieve a comparable result to the state of the art \cite{google2019}, which use about 170 million Wikipedia revision histories with manual modifications. 

\section{Method}
\subsection{GEC system}
\begin{figure*}[t]
	\centering
	\includegraphics[width=1\linewidth,height=0.42\linewidth]{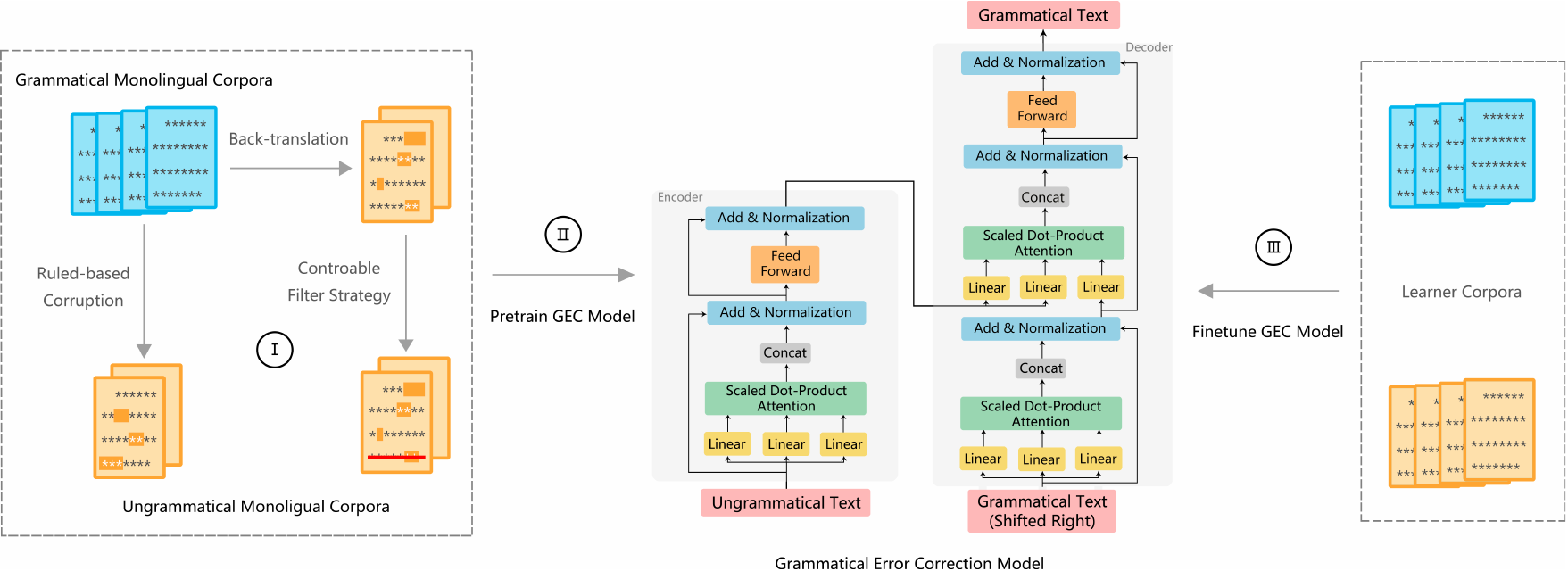}
	\caption{The Grammatical Error Correction system. (\uppercase\expandafter{\romannumeral1}) Introducing errors into grammatical monolingual corpora in two ways. (\uppercase\expandafter{\romannumeral2}) Using synthetic parallel corpora to pretrain the model. (\uppercase\expandafter{\romannumeral3}) Finetuning the GEC model with learner corpora.}
	\label{fig:gecsystem}
\end{figure*}

A lot of recent work has proved the feasibility of using Sequence-to-Sequence Model to solve GEC problems, which showed immense success in neural machine translation. Unlike NMT task, there are not enough available learner corpora in GEC. This motivates the use of data augmentation with monolingual data.

Figure \ref{fig:gecsystem} shows the process of our GEC system. We introduce two important concepts and propose two approaches for data augmentation, including ruled-based method and Back-translation. Furthermore, we put forward a filtering strategy for the corpora to balance error rate and ratio of error types. After pre-training on the synthetic parallel corpora, we fine-tune our GEC model with learner corpora.

\subsection{Error Rate}
Although the task of grammatical error correction is currently regarded as the translation task, the main difference is that both source and target of the GEC use the same language. The original text and its corresponding correction text may be very similar or different. Two examples are shown in the following:

\begin{enumerate}[itemindent=1em,label=(\arabic*),parsep=0.3em,itemsep=0em,leftmargin=0.5em]
	\item Error\ : Students often travel hundreds of \textbf{mile} to get here.
	\item[]Correct\ : Students often travel hundreds of \textbf{miles} to get here.
	\item Error\ : Students often travel \textbf{to here} . 
	\item[]Correct\ : Students often travel \textbf{hundreds of miles to get here} .
\end{enumerate}

The first pair of sentences has few modifications. When the model is over-trained on such data, it is more likely to simply copy the inputs without correcting any words. For the second example, the error sentence contains too little information to be corrected. Models can hardly learn useful information from the training data like this. It is feasible to train decoders with such examples \cite{MonoDataTrainDecoder}, but this is beyond the scope of this work. 

Therefore, we hope to quantitatively analyze the amount of information contained in the corpus. We define the Error Rate ($E_{rate}$) in grammar error correction task as:
\begin{equation}
E_{rate}(S) = \frac{\sum_{i = 0}^n\text{Levenshtein}(S_{i}^{err},S_{i}^{cor})}{\sum_{i = 0}^n\text{Length}(S_{i}^{cor})}
\label{eqa:er}
\end{equation}
where $n$ is the number of sentence pairs in the corpus $S$, $S^{err}$ refers to the source sentence and $S^{cor}$ is the corresponding correction. Levenshtein$(S_{i}^{err},S_{i}^{cor})$ is the shortest edit distance (minimum number of replacements, insertions and deletions) between $S_{i}^{err}$ and $S_{i}^{cor}$ in token level. Length$(S_{i}^{cor})$ denotes the number of tokens in $S_{i}^{cor}$.

\subsection{Error Category}

Another important concept in GEC task is the error types. We use the error categories proposed by Bryant \cite{ErrorType} and further divide them into two kinds of granularity.
\begin{enumerate}[label=(\arabic*),leftmargin=3em]
	\item Coarse-grained error category :
	\begin{itemize}[leftmargin=1em]
		\item Missing type (\textbf{M})
		
		err : w$_0\ \ ...\ \ $w$_{i-1}\ \ \ \ \ \ \ \ $w$_{i+1}\ \ ...\ \ $w$_n$
		
		cor : w$_0\ \ ...\ \ $w$_{i-1}\ \ $\textbf{w}$\bm{_i}\ \ $w$_{i+1}\ \ ...\ \ $w$_n$
		
		Here, \textbf{w}$\bm_i$ is a missing error type.
		\item Unnecessary type (\textbf{U})
		
		err$\ :\ $w$_0\ \ ...\ \ $w$_{i-1}\ \ $\textbf{w}$\bm{_i}\ \ $w$_{i+1}\ \ ...\ \ $w$_n$
		
		cor$\ :\ $w$_0\ \ ...\ \ $w$_{i-1}\ \ \ \ \ \ \ \ $w$_{i+1}\ \ ...\ \ $w$_n$
		
		Here, \textbf{w}$\bm_i$ is an unnecessary error type.
		\item Replacement type (\textbf{R})
		
		err : w$_0\ \ ...\ \ $w$_{i-1}\ \ $\textbf{w}$\bm{_i'}\ \ $w$_{i+1}\ \ ...\ \ $w$_n$
		
		cor : w$_0\ \ ...\ \ $w$_{i-1}\ \ $\textbf{w}$\bm{_i}\ \ $w$_{i+1}\ \ ...\ \ $w$_n$
		
		Here, \textbf{w}$\bm_i$ is a replacement error type.
	\end{itemize}
	
	\item Fine-grained error category :\\
	We distinguish 25 different error types described in \cite{ErrorType}. Such as, \textbf{ADJ} : adjective error, \textbf{ADV} : Adverb error, \textbf{DET} : determiner error and so on.
\end{enumerate}

\subsection{Controllable Rule-based Corpora Corruption}
\begin{table*}[t]
	\centering
	\resizebox{\textwidth}{!}{%
		\begin{tabular}{cp{14cm}}
			\hline
			\multicolumn{1}{c}{\textbf{Error Rate}} & \multicolumn{1}{c}{\textbf{Output}} \\
			\hline
			\multirow{1}{*}{0\%(original)} & Results from dozens of clinical studies will be released at the gathering opening Friday and running through June 2 . \\ 
			\multirow{2}{*}{10\%} & Results from dozens of clinical studies will \textbf{[miss]} released at the gathering opening Friday and \textbf{Overmedicated} running through June 2 . \\ 
			\multirow{2}{*}{30\%} & Results from dozens \textbf{lifestyles} of \textbf{[miss]} studies will be released at \textbf{[miss]} gathering opening Friday and running through \textbf{[miss]} . \\ 
			\multirow{1}{*}{50\%} & Results from \textbf{[miss]} will be \textbf{[miss]} at gathering Friday \textbf{skeptical} and running \textbf{Alyas} through \textbf{alternative} June \textbf{in} 2 \textbf{!} \\ 
			\hline
		\end{tabular}%
	}
	\caption{\label{tab:CorruptExample} Examples of the rule-based corruption with different Error Rates. The \textbf{Bold Font} represents the difference from the original text. \textbf{[miss]} means that the token is deleted.}
\end{table*}
Increasing the size of the training set is a common and effective method to improve performance of neural network. By introducing artificial errors to monolingual corpus, we can get more rich training data for different topic and language phenomenon with the error types we need.

Previous study synthesized data containing determiner and noun number errors and found that the recall on these error types increased significantly, but the recall on other errors declined \cite{thefirstChar-NMT}. Therefore, only increasing partial error types may lead to the deviation of model.

In order to control the error rate and the ratio of error types on synthetic data, we apply the same processing method to each sentence in One Billion Word Benchmark monolingual corpus \cite{OBWBcorpus}. This corpus consist of nearly one billion words of English taken from news articles on the web. We corrupt each token $t$ with a fixed probability $E_{rate}$. The operations of corruption include, deleting the current token $t$ with a probability of $p_m$ (Missing error type), inserting a random token $t'$ in the vocabulary $V$ to the left of $t$ with a probability of $p_u$ (Unnecessary error type) and replacing $t$ with a random token $t'$ in $V$ with a probability of $p_r$ (Replacement error type). Each operation corresponds to an error in this sentence pair. When the replaced token $t \in \mathbb{P}=\{'\ \ "\ \ ,\ \ .\ \ !\ \ ?\}$, it can be only replaced by the elements in $\mathbb{P}$. Algorithm \ref{alg:corruption} formalizes this approach. The vocabulary $V$ is a collection of all different tokens in the monolingual and labeled corpora. 

\begin{algorithm}[h]
	\small
	\caption{Controllable Rule-based Corpora Corruption}
	\label{alg:corruption}
	\KwIn{vocabulary $V$, monolingual corpora $Cm$, target error rate of the synthetic data $E_{rate}$, percentage of the missing error types $p_m$, percentage of the unnecessary error types $p_u$, percentage of the replacement error types $p_r$}
	\KwOut{corruption Corpora $C$}
	Initialize $C$ =\ \{\}\\ 
	\For{all sentences $s$ in $Cm$}{
		\For{ all tokens $t$ in $s$ }{
			rand=Random(1,100)\\
			\If { rand $>$ $E_{rate}$ }{
				continue
			}
			rand=Random(1,100)\\
			\textbf{if } rand $<$ $p_m$ \\
			\ \ \ \ \ \ delete the token $t$ from $s$\\
			\textbf{else if } rand $<$ $p_m+p_u$ \textbf{then}\\
			\ \ \ \ \ \ insert a token in $V$ on the left of $t$\\
			\textbf{else if } rand $<$ $p_m+p_u+p_r$ \textbf{then}\\
			\ \ \ \ \ \ replace $t$ with a token in $V$  \\
			\textbf{end if}		 
		}
		Add $s_i$ to $C$\\
	}
	\Return $C$
\end{algorithm}

Examples of the corruption are shown in Table \ref{tab:CorruptExample}. It shows that the result of corruption is very similar to the original text with the low $E_{rate}$ (10\%). With the increasing of $E_{rate}$, the outputs of corruption become more disorderly.

\subsection{Corpus Generation Based on Back-translation}

For fine-grained error category, it is very difficult to define corrupt rules for all of the 25 error types. This is not only time-consuming, but also difficult to cover all the error phenomena wrote by learners. Therefore, we adopt the Back-translation mechanism for synthesizing error text set. The main idea of this mechanism is to train an error generation model on the corpus of $ correct\ \rightarrow\ error$ sentences, which is the opposite of the error correction model. In the decoding stage, model generates error sentences by introducing noise into monolingual data.

We implement the error generation models in two ways, include phrase based Moses system\footnote{http://www.statmt.org/moses/?n=Moses.Baseline} \cite{Moses} and attention based neural network Transformer \cite{transformer}. Details of the implementation are shown as follows.

\begin{itemize}[leftmargin=1.5em,topsep=0em,parsep=0em,itemsep=0em]
	\item \textbf{Moses-Based error generation model} \\
	We don't use the preprocessing script provided by Moses Open Source Library on the labeled data. We use GIZA++\footnote{https://github.com/moses-smt/giza-pp} for word-aligning our parallel data. The KenLM\footnote{https://kheafield.com/code/kenlm/estimation/} is used for training a trigram language model on the source sentences of the labeled data. The development set is used to fine-tune the model.
	\item \textbf{Transformer-Based error generation model} \\
	We use subword \cite{subword} algorithm to train a Byte Pair Encoding(BPE) vocabulary on labeled data. The labeled and monolingual data is split into sub-words by BPE vocabulary. The union of sub-words is the vocabulary used in error generation model. We evaluate the development set at the end of each epoch and save the model checkpoint with the best cross entropy cost. The settings of hyper parameters are the same as the error correction model (described in section 4.2). We follow the recommendation by Edunov \cite{edunov2018understanding} to restrict sampling over the ten most likely words during inference stage.
\end{itemize}

The proposed error generation models based on Moses and Transformer are trained on Lang8 and NUCLE. The development set is CoNLL2013 test set. Examples of the Back-translation are shown in Table \ref{tab:BTExample}. It can be seen that the decoding results are conservative. Compared with the original sentence, the outputs contain only a few errors. 

\begin{table*}[t]
	\centering
	\resizebox{\textwidth}{!}{%
		\begin{tabular}{cp{14cm}}
			\hline
			\multicolumn{1}{c}{\textbf{Method}} & \multicolumn{1}{c}{\textbf{Output}} \\
			\hline
			\multirow{1}{*}{Original} & Forecasts called for plunging temperatures and afternoon rain storms in the early Southern Hemisphere autumn . \\ 
			\multirow{1}{*}{Moses} & Forecasts called for plunging \textbf{temperature} and afternoon rain \textbf{storm} in the early Southern Hemisphere autumn \textbf{[miss]} \\ 
			\multirow{1}{*}{Transformer} & Forecasts called \textbf{[miss]} plunging temperatures and afternoon rain \textbf{storm} in the early \textbf{South} Hemisphere autumn . \\
			\hline
		\end{tabular}%
	}
	\caption{\label{tab:BTExample} Examples of the Back-translation with different models. The \textbf{Bold Font} represents the difference from the original text. \textbf{[miss]} means that the token is deleted.}
\end{table*}

\subsection{Filtering Strategy for synthetic data }

In the previous section, we find that data synthesized by the Back-translation method contains lower error rate. Training on a corpus with fewer errors will make the models become more conservative when correcting errors. To solve this problem, we propose a filter strategy for the synthetic data. 

The goal of the strategy is to balance error rate and ratio of error types. We greedily remove sentence pairs with lower error rate and those containing error types we do not need. The steps are shown in the following.

\begin{enumerate}[label=\arabic*),topsep=0em,parsep=0em,itemsep=0em,partopsep=0em]
	\item Given the synthetic data $X$ and corresponding corrected data $Y$, target error rate $E_{rate}$, target ratio of $n$ error types $EtRatio=\{\ e_1,\ e_2,\ ...,\ e_n\}$ and the threshold $\theta$ which controls the accuracy of filtering.
	\item Calculate the error rate of each sentence pair and arrange them in ascending order.
	\item Remove sentence pairs greedily until error rate of $E_{rate}*(1-\theta)$ is achieved.
	\item Calculate the $n$ error types for each sentence pairs\footnote{https://github.com/chrisjbryant/errant/blob/master/parallel \_to\_m2.py} and count them as $Et=[\ num_1,\ num_2,\ ...,\ num_n]$.
	\item Count the sum of $n$ error types in the synthetic data and record them as $EtSum=\{\ sum_1,\ sum_2,\ ...,\ sum_n\}$.
	\item Find the subscript $i$ for the lowest target error type ratio $EtRatio[i]$, mark the number $EtSum[i]$ as benchmark $B$. The benchmark is the number of output for each proportion.
	\item Using algorithm \ref{alg:FilterStrategy} to filter out sentence pairs that contain error types beyond the target ratio.
\end{enumerate}

\begin{algorithm}[h]
	\small
	\caption{Filter Strategy for balancing the ratio of error types}
	\label{alg:FilterStrategy}
	\KwIn{the error-corrected pairs and the corresponding number of $n$ error types $Array=\{x,\ y,\ Et=[num_1,num_2,$ $\ ...,\ num_n]\}$, target ratio of $n$ error types $EtRatio$ = $\{\ e_1,\ e_2,$ $\ ...,\ e_n\}$, threshold $\theta$, benchmark $B$, the sum of $n$ error types $EtSum=[\ sum_1,$ $\ sum_2,\ ...,\ sum_n]$}
	\KwOut{Filtered synthetic data $Array$}
	\For{all sentence pairs $a$ in $Array$}{
		\If{all error types $Et[i]$ in $a.Et$ meet the \\
			$\ \ \ \ \ \ $condition: $EtSum[i]$ - $Et[i]$ $<$ \\
			$\ \ \ \ \ \ $EtRatio[i]* $B$ * ( 1 - $\theta$ )}{
			delete $a$ from $Array$\;
			update $EtSum$\;
		}
		\If{all error types $Et[i]$ in $EtSum$ meet the \\
			$\ \ \ \ \ \ $condition : $Et[i]$ $>$ $EtRatio[i]$ *$B$* \\
			$\ \ \ \ \ \ $(1 + $\theta$)}{
			break\;
		}
	}
	\Return Array
\end{algorithm}

\section{Experiment Setup}
\subsection{Datasets}

We use NUCLE \cite{NUCLEcorpus} and Lang8 \cite{Lang-8} as our labeled training data. We choose the test set from the CoNLL 2013 shared task \cite{CoNLL2013} as our development set and report MaxMatch(M2) score \cite{Maxmatch} on the CoNLL-2014 shared task \cite{CoNLL2014} test set. 
The monolingual corpus we use is the One Billion Word Benchmark \cite{OBWBcorpus}.

We remove the uncorrected sentence pairs from the labeled data like previous work \cite{yuanfudao}. By analyzing the data, we find that there are many instances containing URLs in NUCLE, illegal characters and emoji in Lang8. We remove the sentence pairs containing these cases from all training sets. The statistics of the corpora after preprocessing are given in Table \ref{tab:corpus}.

\begin{table}[h]
	\begin{center}
		\begin{tabular}{ccc}
			\hline
			\textbf{Corpus} & \textbf{\#sentences} & \textbf{\#tokens} \\
			\hline 
			NUCLE & 21,177 & 0.5M \\
			Lang-8 & 1,230,095 & 18.5M \\
			One-Billion* & 19,966,137 & 497M \\
			CoNLL2013 & 1,382 & - \\
			CoNLL2014 & 1,312 & -\\
			\hline
		\end{tabular}
	\end{center}
	\caption{\label{tab:corpus}  Statistics for the sentence pairs of all data after preprocessing. Here, * refers that this corpus is unlabeled. }
\end{table}

\subsection{Hyper Parameters}

In this work, we use the Transformer model \cite{transformer} implemented by FAIR\footnote{https://github.com/pytorch/fairseq} tool kits as the error correction model.

The detailed parameters of the model are as follows: Both of the source embedding and the target embeddings have 512 dimensions and use the same BPE vocabulary. We share the weights of decoder's input and output embeddings. Both of the encoder and decoder have 6 multi-head layers and 8 attention heads. The size of inner layer at each multi-head layer is 2048. We use Adam optimizer to train transformer model with the inverse squared root schedule which will decay the learning rate based on the inverse square root of the warm-up steps. The learning rate initialize with $5 \times 10^{-4}$ and warm-up during the first 4,000 steps. In order to train transformer adequately, we use a batch size of 32,000 tokens and fine-tune the model on labeled data for 30,000 steps. Dropout is applied at a ratio of 0.3. The Loss function we use is the Edit-weighted MLE objective \cite{JD2018b} and the factor $\Lambda$ is set to 1.2.

We use a random seed value of 3 for training process and save the model checkpoints for each epoch. The ensemble model consists of eight checkpoints that get the best loss value on development set during fine-tuning. Calculate the geometric average \cite{google2019} of all single model outputs as the final output of the ensemble model.
During inference stage, we set beam size to 12 and keep the best decoding results as the model output without using any re-ranking.

\section{Experiment and Analysis}
As mentioned in Junczys-Dowmunt, Sequence-to-Sequence training of neural networks is a recognition optimization process, so it is prone to instability \cite{JD2018b}. We use different random seeds to initialize the model. Table \ref{tab:instability} shows that different initialization weights have a great influence on the performance of the model. Therefore, we will report the average of the model scores initialized by random seed=1, 2 and 3 (not ensemble) in all exploratory experiments.

\begin{table}[h]
	\centering
	\begin{tabular}{cccc}
		\hline
		\textbf{Seed} & \textbf{Precision} & \textbf{Recall} & \textbf{F}$\bm{_{0.5}}$ \\ \hline
		1             & 57.44              & 32.25           & 49.68         \\ 
		2             & 55.56              & 30.80           & 47.86         \\ 
		3             & 56.67              & 32.19           & 49.19         \\ \hline
	\end{tabular}
	\caption{\label{tab:instability}  Scores of models initialized with different random seeds.}
\end{table}

We will analyze the impact of different error rates and ratio of error types on synthetic data and also verify the effectiveness of the filtering strategy. We use the ERRANT\footnote{https://github.com/chrisjbryant/errant} to evaluate the model performance by different error types.

\subsection{Performance impact by different Error Rates}

Firstly, we investigate the impact of different error rates by the synthetic data on model performance. We corrupt 1.25 million (1:1 to the labeled data) monolingual corpus by rule-based method described in Section 3.3. The ratio of the insertion, deletion and replacement operations is 1:1:1.

\begin{figure}[h]
	\centering
	\includegraphics[width=1\linewidth,height=0.32\linewidth]{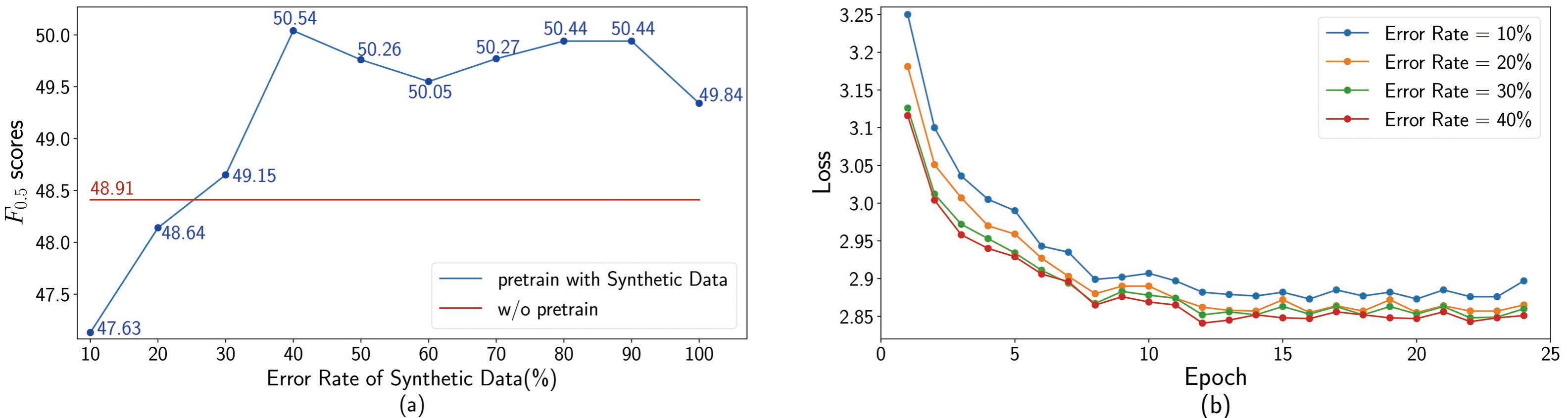}
	\caption{(a) F$_{0.5}$ scores of models pre-training on the synthetic data with different error rates corrupted by rule-based method. The red line is the baseline model training on the labeled data. The blue line is the models pre-training on the different synthetic data and fine-tuning with the labeled data. (b) The loss value variation of four fine-tuning models which are pre-trained on synthetic data with different error rate.}
	\label{fig:ErrorRate}
\end{figure}

We generate 10 sets of synthetic data with different error rates range from 10\% to 100\%. Each set of the synthetic data is trained for five epochs. Then, the labeled data is performed 30K updates during fine-tuning. The other parameters of the model used in the pre-training stage are the same as those in the fine-tuning stage (described in Section 4.2). We record the scores of three models initialized with random seed=1, 2, 3 and report the average of them (The blue line in Figure \ref{fig:ErrorRate}(a)). We also train a baseline model that only use the labeled data without pre-training (The red line in Figure \ref{fig:ErrorRate}(a)).

Figure \ref{fig:ErrorRate}(a) shows that pre-training on the synthetic data with error rate less than 30\% is unhelpful to improve the performance of the error corrected model. When training on the synthetic data with few mistakes ($<=$10\%), the performance is declined greatly compared with the baseline. This also confirms what Xie said, "the addition of examples where source and target are identical data may also cause the model to become too conservative with edits and thus reduce the recall of the system" \cite{xie2018noising}. Therefore, we should avoid using synthetic data with low error rate, even labeled data. 

The model performance gets best when error rate is 40\%.
With the increasing of error rate ($>=40\%$) on synthetic data, the scores of the model do not change significantly. However, it is worth noting that, pre-training with the high error rate ($>=80\%$) is still helpful to improve the model performance. Theoretically, the source of these data should be extremely disorderly. But experiments show that the error corrected model can still learn useful information from the noisy source.

Figure \ref{fig:ErrorRate}(b) shows the variation of loss value for models pretraining on synthetic data with different error rate. The starting point of the four lines represents the superiority of the pre-training results for synthetic data with different error rate. Pre-training GEC model using synthetic corpus with error rate of 40\% (red line in (b)) can make the model initialized into a more reasonable parameter space. The effect of pre-training still has a great impact on the process of fine-tuning, since we can see that the red line (Error Rate = 40\%) has been kept at a very low value in the Figure \ref{fig:ErrorRate}(b).

\subsection{Performance impact by different ratio of Error Types}

To make a comparison with the previous experiments, we still generate 1.25 million data with the rule-based methods and set $E_{rate}$ to $30\%$. We vary the proportion of the three operations, $p_m=20\%$, $p_u=20\%$,  $p_r=60\%$ as the first group of experiments recorded as 1:1:3. The other two groups are, $p_m=60\%$, $p_u=20\%$, $p_r=20\%$ and $p_m=20\%$, $p_u=60\%$, $p_r=20\%$ denoted as 3:1:1 and 1:3:1, respectively. The procedure of pre-training synthetic data is the same as before. 

\begin{figure}[h]
	\centering
	\includegraphics[width=0.9\linewidth,height=0.55\linewidth]{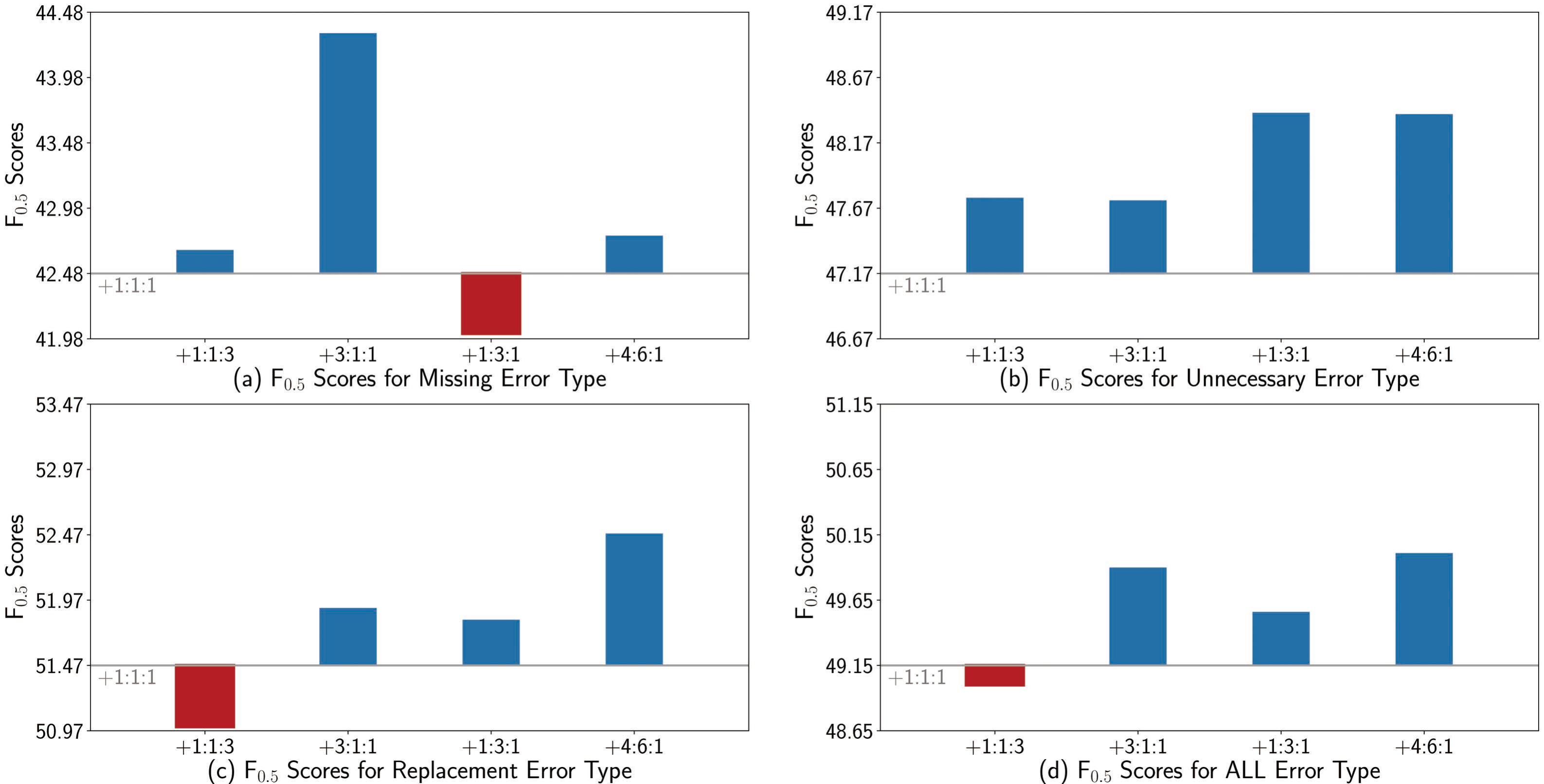}
	\caption{Scores of models pre-training on the synthetic data corrupted by rule-based method with different ratio of error types. (a), (b), (c) and (d) represent the F$_{0.5}$ scores for Missing, Unnecessary, Replacement and ALL error type, respectively.}
	\label{fig:ErrorType}
\end{figure}

Figure \ref{fig:ErrorType} summarizes the results about these experiments. We separately evaluate the F$_{0.5}$ score of coarse-grained error category, Missing(M), Unnecessary(U) and Replacement(R). The performance of 3:1:1 group, which has more missing type errors, shows a significant increase in M type compared with 1:1:1. The group of 1:3:1 performs well in U type and it is as expected.

Unexpectedly, the experiment of 1:1:3 gets the worst result and does not improve in R type. The model may not understand the replacement operations very well. In terms of the results, adding more replacement errors of synthetic data just makes the imodel learn better on missing and unnecessary type errors. Because a replacement operation consists of an insert and a delete operation.

The ratio of M:U:R in the original training set is 29:12:59 and the error rate is approximate 34\%. We generate a set of data with $E_{rate}=30\%$ $p_m=36\%$ $p_u=55\%$ $p_r=9\%$(4:6:1), which makes the M:U:R on the total data to 1:1:1. The model achieves the best score 50.01.

\subsection{Performance impact by different size of Synthetic Data}

\begin{figure}[h]
	\centering
	\includegraphics[width=0.8\linewidth,height=0.55\linewidth]{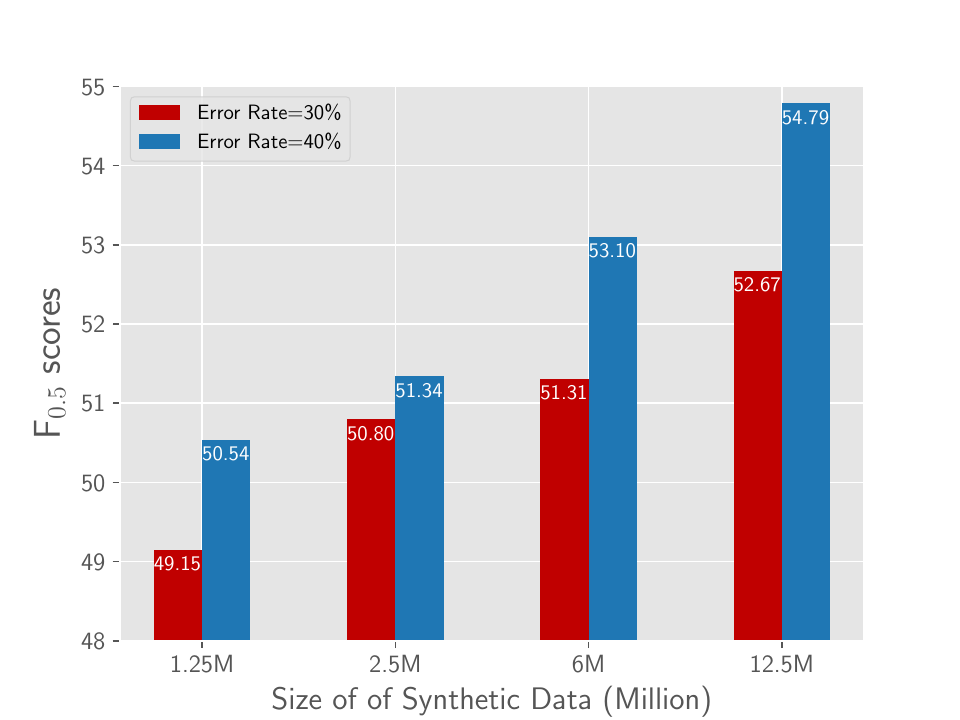}
	\caption{F$_{0.5}$ scores of models pre-training on the synthetic data with different number of synthetic data. Each error rate contains four sets of experiments, including  1:1 (1.25M), 2:1 (2.5M), 4:1 (6M) and 10:1 (12.5M).}
	\label{fig:Quantity}
\end{figure}

In order to investigate whether performance improvement is increasing with higher ratios of synthetic data, we perform four experiments for different error rate of synthetic data.

Figure \ref{fig:Quantity} shows that when the size of the synthetic data is the same, the corpus with 40\% error rate contributes more to the model than the corpus with 30\% error rate. As the scale of synthetic data increasing, the performance of the model becomes better and better. This conflicts with the conclusion of Fadaee et al. \cite{fadaee2018back} in machine translation task. They find that translation quality does not improve linearly with the size of the synthetic data. The model trained on (1:4) real to synthetic ratio of training data achieves the best results.

In our opinion, the main reason for the different conclusions is the particularity of grammatical error correction task. Since the source and target language is the same in GEC, we can regard it as a denoising procedure. The pseudo-data generated by adding noise to rich contexts can be used to improve the denoising ability of the model.

\subsection{Effectiveness of the Filtering Strategy}

The goal of the filtering strategy is to control error rate and ratio of error types in corpus. In previous experiments, we have found that adjusting the error rate or error types of synthetic corpus has a great influence on model performance. Therefore, we further verify that the filtering strategy for fine-grained error types is also effective for data generated by Back-translation.

\begin{figure}[h]
	\centering
	\includegraphics[width=0.6\linewidth,height=0.40\linewidth]{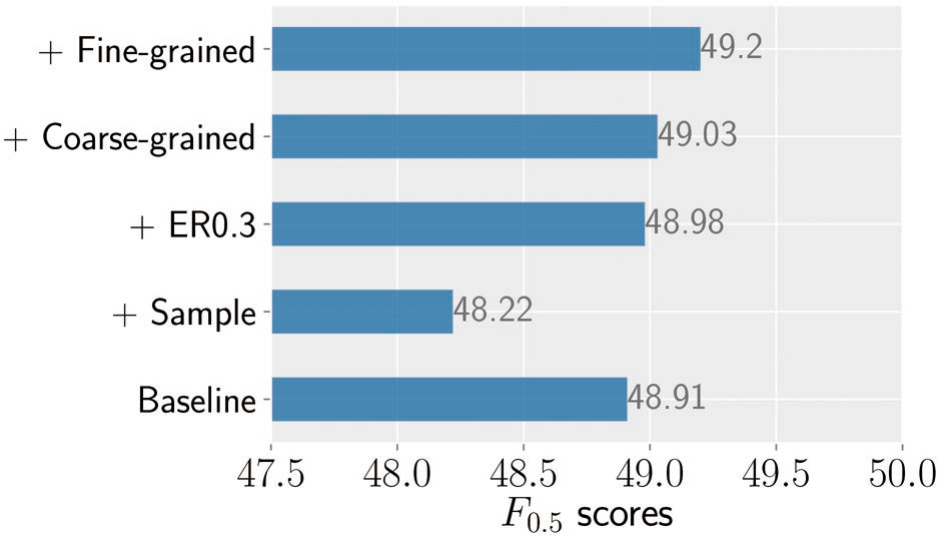}
	\caption{Scores of models pre-training on the synthetic data generated by Back-translation. $Baseline$ refers the model training on labeled data without pre-training. }
	\label{fig:backtranslation}
\end{figure}

The two methods described in Section 3.4 is used to introduce noise to the monolingual data. We randomly sample 1.25 million lines of the union of two output sets ($+ sample$). We filter out sentence pairs in the two output sets with low error rate to make the remaining corpus $E_{rate}$ = 30\% and randomly sample 1.25 million lines ($+ ER0.3$). The algorithm \ref{alg:FilterStrategy} is applied on the remaining corpus, making the ratio of coarse-grained error category 1:1:1($+ Coarse$-$gained$) or the ratio of fine-grained error category 1:1:1:...:1($+ Fine$-$gained$).

As it can be seen in Figure \ref{fig:backtranslation}, the performance of group $+sample$ is much worse than that of group $Baseline$. This indicates that using the unfiltered data generated by Back-translation is unhelpful for the training of the model. After the sentence pairs with low error rate are filtered out ($+ER0.3$), the model score is improved by 0.76 than $+sample$. Both the result of $+Coarse$-$gained$ and $+ Fine$-$gained$ show that the filtered strategy can further improve the score of the model by balancing the proportion of error types in the corpus. 

Previous experiments have shown that increasing the number of partial error types leads the model being more inclined to modify specific types of errors, which is the reason for the poor result of group $+Sample$. The most important contribution of filtering strategy is to keep the same proportion of all error types as possible to improve the performance of the model.

\subsection{Comparison with Other systems}

We use the back-translation method to decode 100M monolingual data. After filtering out the unqualified data in the synthetic corpus, there was about 10M data left. Then, we synthesize 90M parallel data using rule-based method. All synthetic corpora have about 40\% error rate and the same proportion of error types. In order to generate enough data, we use the monolingual corpus published by Junczys-Dowmunt \cite{junczys2016phrase}. Neither method of synthesizing data is aimed at spelling errors, so we add a spell checker based on CyHunspell\footnote{https://pypi.org/project/CyHunspell/} with a language model training on the monolingual data to correct the spelling errors. 

\begin{table*}[t]
	\centering
	\resizebox{\textwidth}{!}{
		\begin{tabular}{ccccc}
			\hline
			\multirow{2}{*}{\textbf{Model}} & \multirow{2}{*}{\textbf{Learner Corpora}} & \multicolumn{3}{c}{\textbf{CoNLL-2014}} \\
			&  & \textbf{Precision} & \textbf{Recall} & \textbf{F$_{0.5}$} \\ \hline 
			\multirow{2}{*}{CNN Seq2Seq(4 ens.+LM+spellcheck) \cite{NUS2018CNN}} & \multirow{2}{*}{L8,NUCLE} & \multirow{2}{*}{65.49} & \multirow{2}{*}{33.14} & \multirow{2}{*}{54.79} \\
			&  &  &  &  \\ 
			\multirow{2}{*}{Transformer(4 ens.+Many training strategies) \cite{JD2018b}} & \multirow{2}{*}{L8,NUCLE} & \multirow{2}{*}{63.00} & \multirow{2}{*}{38.90} & \multirow{2}{*}{56.10} \\
			&  &  &  &  \\ 
			\multirow{2}{*}{SMT + BiGRU(4 ens.+LM+spellcheck) \cite{JD2018a}} & \multirow{2}{*}{L8,NUCLE} & \multirow{2}{*}{66.77} & \multirow{2}{*}{34.49} & \multirow{2}{*}{56.25} \\
			&  &  &  &  \\ 
			\multirow{2}{*}{CNN Seq2Seq + Quality Estimation(12 ens.) \cite{NUS2018Quality}} & \multirow{2}{*}{L8,NUCLE,FCE} & \multirow{2}{*}{-} & \multirow{2}{*}{-} & \multirow{2}{*}{56.52} \\
			&  &  &  &  \\ 
			\multirow{2}{*}{Data Generation + Transformer(8 ens.) \cite{google2019}} & \multirow{2}{*}{L8,NUCLE} & \multirow{2}{*}{66.70} & \multirow{2}{*}{43.90} & \multirow{2}{*}{60.40} \\
			&  &  &  &  \\ 
			\multirow{2}{*}{Ours(8 ens.)} & \multirow{2}{*}{L8,NUCLE} & \multirow{2}{*}{\textbf{70.63}} & \multirow{2}{*}{37.04} & \multirow{2}{*}{\textbf{59.79}} \\
			&  &  &  &  \\ \hline
		\end{tabular}
	}
	\caption{Comparison of state-of-the-art GEC systems on CoNLL-2014 test set. Our ensemble model is a geometric average of eight best checkpoints for a single model. Here, - represents that the result is not public.}
	\label{tab:result}
\end{table*}

The performance with the well known GEC systems as shown in Table \ref{tab:result}. ``Data Generation + Transformer" uses about 170 million Wikipedia revision histories with manual modification to pre-train the model. Pre-training the model with synthetic data makes our performance surpass most systems. We only use fewer but more accessible synthetic data to achieve result that is comparable to Lichtarge \cite{google2019}.

\section{Discussion}

We performe ablation analysis on our grammatical error correction system. The results of each parts are shown in Table \ref{tab:ablation}.

\begin{table}[h]
	\small
	\centering
	\begin{tabular}{lccc}
		\hline
		\textbf{\footnotesize Model} & \textbf{\footnotesize P} & \textbf{\footnotesize R} & \textbf{\footnotesize F$_{0.5}$} \\ \hline
		Transformer  & 61.25 & 33.52 & 52.56 \\
		\ \ + 100M   & 66.64 & 32.65 & 55.16 \\
		\ \ \ \ + SpellChecker  & 67.98 & 34.55 & 56.96 \\
		\ \ \ \ \ \ + 8 ens.  & 70.63 & 37.04 & 59.79 \\ \hline
	\end{tabular}%
	\caption{ Results on the CoNLL-2014 test sets. + 100M means using the 100M synthetic data. + SpellChecker refers to add the spelling checker based on the language model mentioned above.}
	\label{tab:ablation}
\end{table}

Only using the Transformer model can make the GEC system perform well. Further use of both large-scale monolingual corpus for data augmentation and model ensemble can greatly improve the performance of the system. Because of the input is token level, Transformer doesn't pay much attention to the information between characters. This allows the spell checker to achieve improvement.

\begin{figure}[h]
	\centering
	\includegraphics[width=1\linewidth,height=0.8\linewidth]{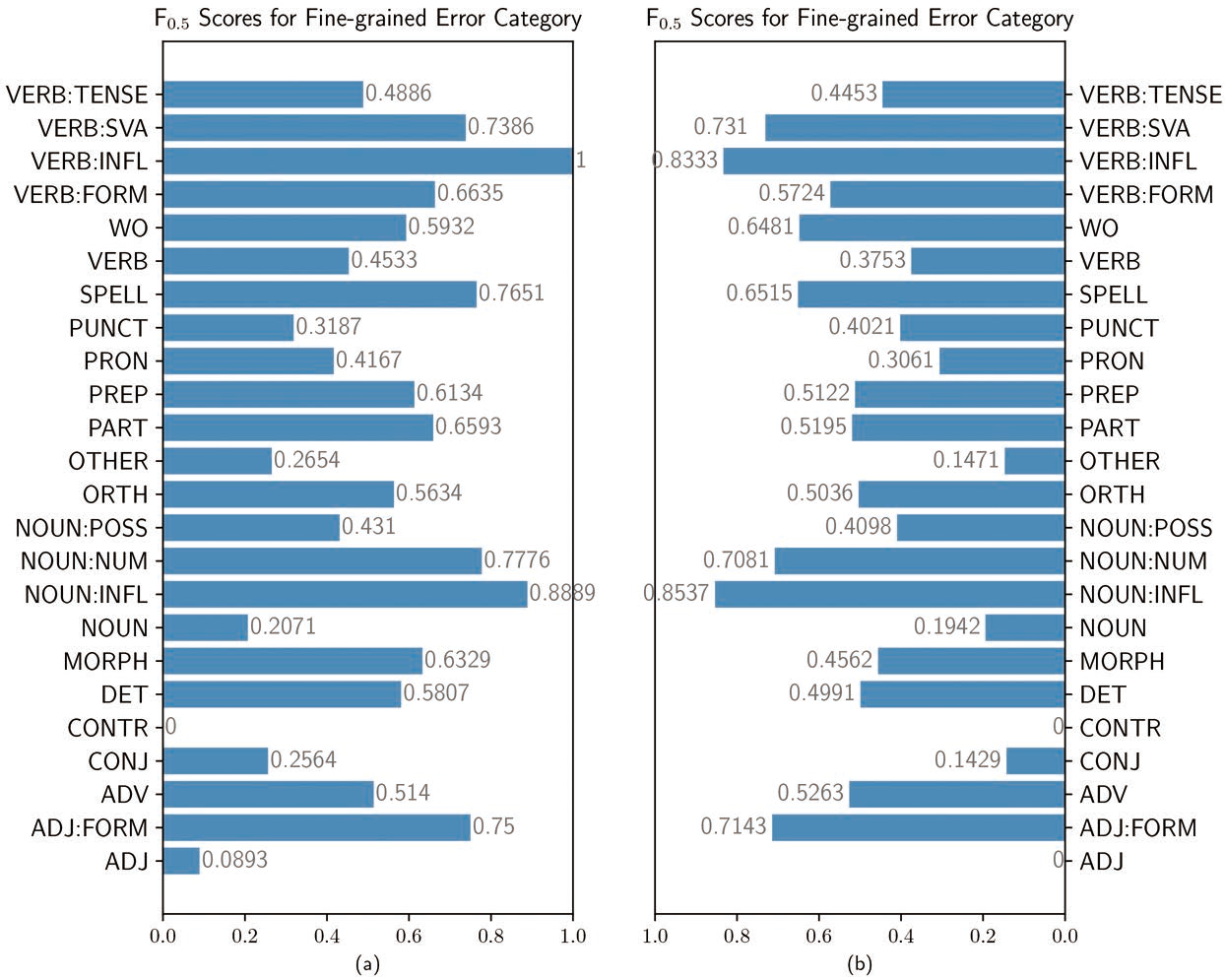}
	\caption{F$_{0.5}$ scores for Fine-grained Error Category. (a) The model pre-training on the 100M synthetic data. (b) without pre-training }
	\label{fig:dawithnoda}
\end{figure}

We use the open source tool ERRANT \cite{ErrorType} to analyze the F$_{0.5}$ score of the two models for correcting different types of errors. The results of using pre-training data or not are shown in Figure \ref{fig:dawithnoda}, respectively. Comparing the two groups of experiments, we can see that the model pre-training with synthetic data has better performance in most error types than that without using synthetic data. Especially in Verb Inflection (VERB:INFL), Pronoun (PRON), Preposition (PREP), Morphology (MORPH) and Conjunction (CONJ) errors, scores increased by more than 10 percent. However, the model performance declines 5 percent in terms of word order (WO) errors. This may be due to the fact that our data synthesis method does not produce more word order errors. In future experiments, we will consider using the method of disrupting the order of sentences.

\section{Related Work}
Felice et al. propose a system that combines a rule-based method and SMT system augmented by a language model \cite{CAMB_CoNLL2014}. Junczys-Dowmunt et al. \cite{AMU_CoNLL2014} use large-scale Lang-8 corpus to tune the Moses system \cite{Moses}. These two studies prove that SMT system can combine large corpus and perform well in GEC task. To address the problem of lacking granularity below the word level, Chollampatt  uses a character-level SMT to correct the misspelled words \cite{chollampatt2017connecting}.

The early work of neural machine translation can not exceed the statistical machine translation at the same period for GEC task \cite{thefirstNMTinGEC,thefirstChar-NMT}.
Chollampatt et al. use a multilayer convolutional encoder-decoder neural network which outperforms the current state-of-the-art statistical machine translation-based approach \cite{NUS2018CNN}. Grundkiewicz et al. believe that statistical machine translation could not be replaced by deep neural networks. They combine the RNN with a phrase-based SMT system to achieve a higher score \cite{JD2018a}. Junczys-Dowmunt et al. apply Transformer \cite{transformer} to the GEC task and achieves good performance \cite{JD2018b}. 

Motivated by the data scarcity for the GEC task, many researchers focus on generating corpus for data augmentation. Some works use rule-based methods to generate artificial error data that contains specific types of errors \cite{yuan2013constrained,thefirstChar-NMT,rei2017artificial}. Rozovskaya et al. construct confusion sets to replace specific words in error-free text. There are fewer types of errors covered in synthetic corpus \cite{rozovskaya2010generating}. Felice et al. use linguistic information to characterize contexts of naturally occurring errors and replicate them in error-free text \cite{felice2014generating}. Neither of their methods takes full account of the impact of different error rates and error type ratios on the results. Ge et al. propose a dual reinforcement learning strategy that alternately train grammatical error correction and generation model, without using monolingual data \cite{ge2018fluency}. Unlike them, we use the generation model to decode the monolingual corpus and synthesize a large number of additional grammatical error correction data. Both Rei \cite{rei2017artificial} and Xie \cite{xie2018noising} add noise to monolingual data using Back-translation mechanism of SMT and NMT, respectively. Greedy use of all synthetic corpus makes the models fail to achieve a high score. Lichtarge et al. extract the edits between snapshots from the Wikipedia revisions history to generate parallel corpus \cite{google2019}.

\section{Conclusion}
To solve the Grammatical Error Correction problem, we use Transformer which is a powerful deep sequence-to-sequence architecture to translate ungrammatical text into right one.
Motivated by the data scarcity for the GEC, we introduce two important concepts, error rate and error type, and propose two data synthesis methods that can control the error rate and ratio of error types on the synthetic data. The first approach is to use different proportions of insert, delete and replace operations to corrupt monolingual corpus. The second approach is to introduce noise into the monolingual corpora by using the Back-translation mechanism and use a controllable method to filter the synthetic parallel data.
Both rule-based and Back-translation methods can be easily applied to any domain or language of monolingual corpora (e.g., biomedical domain, German).

By controlling variables, we validate the effects of different error rates and ratio of error types on the model performance. When the error rate is 40\% and the proportion of different error types is the same, the performance of the model can improve significantly. Our method does not depend on any external knowledge and manual modification, which can achieve comparable score as the state of the art.

Correcting misuse of nouns (NOUN) and verbs (VERB) remains a challenging issue, which requires the understanding of the context and the ability of inference. In the future work, we will try to add some reasoning mechanisms to the sequence-to-sequence model. Grammatical corpus will be used to synthesize more word order errors in order to improve the performance on correcting WO errors.

\section*{Acknowledgment}

This work is supported by the National Key R\&D Program of China under grant No.2018YFC1900804 and Research Program of State Language Commission under grant No.YB135-89.
\end{CJK*}

\bibliographystyle{acl_natbib}
\bibliography{acl2019}
\end{document}